\pdfoutput=1

\documentclass[11pt]{article}

\usepackage[]{ACL2023}

\usepackage{times}
\usepackage{latexsym}

\usepackage{graphicx} 
\usepackage{amsmath}
\usepackage{amsthm}
\usepackage{booktabs}
\usepackage[ruled,linesnumbered]{algorithm2e}
\usepackage{color}
\usepackage{multirow}
\usepackage{array}
\usepackage{bbm}
\usepackage{arydshln}
\usepackage{dsfont}
\usepackage[utf8]{inputenc}
\usepackage{cleveref}
\crefname{section}{§}{§§}
\Crefname{section}{§}{§§}
\usepackage{makecell}
\usepackage{bbding}
\usepackage{amssymb}
\usepackage{pifont}%

\usepackage[T1]{fontenc}

\usepackage[utf8]{inputenc}

\usepackage{microtype}

\usepackage{inconsolata}

\title{DiffusionNER: Boundary Diffusion for Named Entity Recognition}

\author{
Yongliang Shen$^{1 \ast}$, Kaitao Song$^{2\dagger}$, Xu Tan$^2$,\\
\textbf{Dongsheng Li$^2$, Weiming Lu$^{1\dagger}$, Yueting Zhuang$^1$}\\
  Zhejiang University$^1$, Microsoft Research Asia$^2$ \\ 
  \texttt{\{syl, luwm\}@zju.edu.cn},
  \texttt{\{kaitaosong, xuta\}@microsoft.com}
}

\begin{document}
\maketitle

\renewcommand{\thefootnote}{\fnsymbol{footnote}}
\footnotetext[1]{\;This work was done when the first author was an intern at Microsoft Research Asia.}
\footnotetext[2]{\;Corresponding author.}
\renewcommand{\thefootnote}{\arabic{footnote}}

\begin{abstract}

In this paper, we propose \textsc{DiffusionNER}, which formulates the named entity recognition task as a boundary-denoising diffusion process and thus generates named entities from noisy spans. During training, \textsc{DiffusionNER} gradually adds noises to the gold entity boundaries by a fixed forward diffusion process and learns a reverse diffusion process to recover the entity boundaries. In inference, \textsc{DiffusionNER} first randomly samples some noisy spans from a standard Gaussian distribution and then generates the named entities by denoising them with the learned reverse diffusion process. 
The proposed boundary-denoising diffusion process allows progressive refinement and dynamic sampling of entities, empowering \textsc{DiffusionNER} with efficient and flexible entity generation capability.
Experiments on multiple flat and nested NER datasets demonstrate that \textsc{DiffusionNER} achieves comparable or even better performance than previous state-of-the-art models\footnote{\;Our code will be available at \url{https://github.com/tricktreat/DiffusionNER}.}.

\end{abstract}

\section{Introduction}

Named Entity Recognition (NER) is a basic task of information extraction \citep{tjong-kim-sang-de-meulder-2003-introduction}, which aims to locate entity mentions and label specific entity types such as {person}, {location}, and {organization}. It is fundamental to many structured information extraction tasks,
such as relation extraction \citep{li-ji-2014-incremental, miwa-bansal-2016-end} and event extraction \citep{mcclosky-etal-2011-event, wadden-etal-2019-entity}.

\begin{figure}
    \centering
    \includegraphics[width=\linewidth]{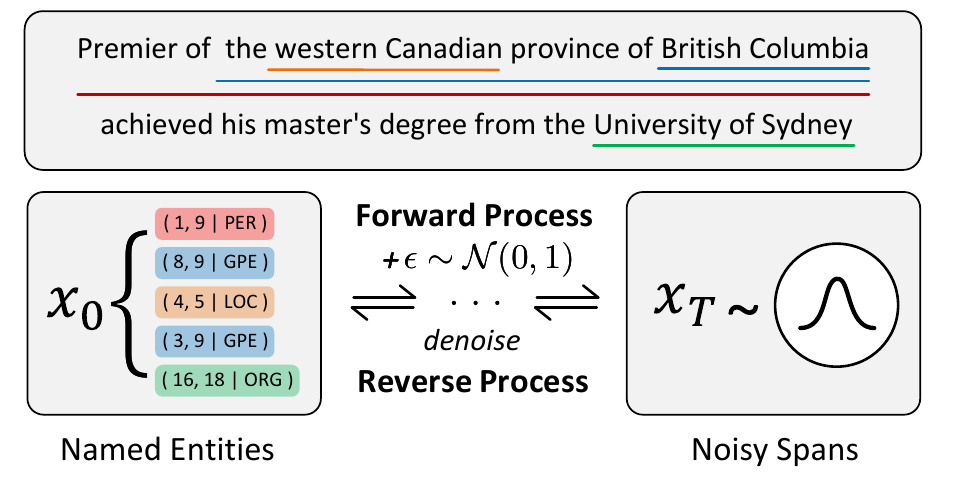}
    \caption{Boundary diffusion in named entity recognition. The fixed forward diffusion process adds Gaussian noise to the entity boundaries at each timestep, and the noisy boundaries recover original state by denoising with the learnable reverse diffusion process. For inference, the reverse diffusion process generates entity boundaries and performs entity typing based on the noisy spans sampled from the  Gaussian distribution.}
    \label{fig:intro}
\end{figure}

Most traditional methods \citep{10.1162/tacl_a_00104} formulate the NER task into a sequence labeling task by assigning a single label to each token.
To accommodate the nested structure between entities, some methods \citep{ju-etal-2018-neural, wang-etal-2020-pyramid} further devise cascaded or stacked tagging strategies.
Another class of methods treat  NER as a classification task on text spans \citep{sohrab-miwa-2018-deep, Eberts}, and assign labels to word pairs \citep{yu-etal-2020-named, li2022unified} or potential spans \citep{lin-etal-2019-sequence, shen-etal-2021-locate}. In contrast to the above works, some pioneer works \citep{paolini2021structured, yan-etal-2021-unified-generative, lu-etal-2022-unified} propose generative NER methods that formulate NER as a sequence generation task by translating structured entities into a linearized text sequence. However, due to the autoregressive manner, the generation-based methods suffer from inefficient decoding. In addition, the discrepancy between training and evaluation leads to exposure bias that impairs the model performance.

We move to another powerful generative model for NER, namely the diffusion model. As a class of deep latent generative models, diffusion models have achieved impressive results on image, audio and text generation \citep{rombach2021highresolution, dalle2, kong2021diffwave, Li-2022-DiffusionLM, gong2022diffuseq}. The core idea of diffusion models is to systematically perturb the data through a forward diffusion process, and then recover the data by learning a reverse diffusion process.

Inspired by this, we present \textsc{DiffusionNER}, a new generative framework for named entity recognition, which formulates NER as a denoising diffusion process \citep{pmlr-v37-sohl-dickstein15, ddpm} on entity boundaries and generates entities from noisy spans. As shown in Figure \ref{fig:intro}, during training, we add Gaussian noise to the entity boundaries step by step in the forward diffusion process, and the noisy spans are progressively denoised by a reverse diffusion process to recover the original entity boundaries. The forward process is fixed and determined by the variance schedule of the Gaussian Markov chains, while the reverse process requires learning a denoising network that progressively refines the entity boundaries. For inference, we first sample noisy spans from a prior Gaussian distribution and then generate entity boundaries using the learned reverse diffusion process.

Empowered by the diffusion model, \textsc{DiffusionNER} presents three advantages. 
First, the iterative denoising process of the diffusion model gives \textsc{DiffusionNER} the ability to progressively refine the entity boundaries, thus improve performance. 
Second, independent of the predefined number of noisy spans in the training stage, \textsc{DiffusionNER} can sample a different number of noisy spans to decode entities during evaluation. Such dynamic entity sampling makes more sense in real scenarios where the number of entities is arbitrary. Third, different from the autoregressive manner in generation-based methods, \textsc{DiffusionNER} can generate all entities in parallel within several denoising timesteps. 
In addition, the shared encoder across timesteps can further speed up inference. 
We will further analyze these advantages of \textsc{DiffusionNER} in \Cref{ana}. In summary, our main contributions are as follows:

\begin{itemize}
    \item \textsc{DiffusionNER} is the first to use the diffusion model for NER, an extractive task on discrete text sequences. Our exploration provides a new perspective on diffusion models in natural language understanding tasks.
    \item \textsc{DiffusionNER} formulates named entity recognition as a boundary denoising diffusion process from the noisy spans. \textsc{DiffusionNER} is a novel generative NER method that generates entities by progressive boundary refinement over the noisy spans.
    \item We conduct experiments on both \textit{nested} and \textit{flat} NER to show the generality of \textsc{DiffusionNER}. Experimental results show that our model achieves better or competitive performance against the previous SOTA models.
\end{itemize}

\section{Related Work}

\subsection{Named Entity Recognition}
Named entity recognition is a long-standing study in natural language processing. Traditional methods can be divided into two folders: tagging-based and span-based. For tagging-based methods~\citep{10.1162/tacl_a_00104, ju-etal-2018-neural, wang-etal-2020-pyramid}, they usually perform sequence labeling at the token level and then translate into predictions at the span level. Meanwhile, the span-based methods~\citep{sohrab-miwa-2018-deep, Eberts, shen-etal-2021-locate, li2022unified} directly perform entity classification on potential spans for prediction. Besides, some methods attempt to formulate NER as sequence-to-set~\citep{ijcai2021-542, ijcai2022p613} or reading comprehension~\citep{li-etal-2020-unified, shen-etal-2022-parallel} tasks for prediction.
In addition, autoregressive generative NER works~\citep{athiwaratkun-etal-2020-augmented, decao2021autoregressive, yan-etal-2021-unified-generative, lu-etal-2022-unified} linearize structured named entities into a sequence, relying on sequence-to-sequence language models, such as BART~\citep{lewis-etal-2020-bart}, T5~\citep{2020t5}, etc., to decode entities. These works designed various translation schemas, including from word index sequence to entities \citep{yan-etal-2021-unified-generative} and from label-enhanced sequence to entities \citep{paolini2021structured}, to unify NER to the text generation task and achieved promising performance and generalizability. Other works \citep{zhang-etal-2022-de} focus on the disorder of the entities and mitigate incorrect decoding bias from a causal inference perspective.

Different from previous works, our proposed \textsc{DiffusionNER} is the first one to explore the utilization of the generative diffusion model on NER, which enables progressive refinement and dynamic sampling of entities. Furthermore, compared with previous generation-based methods, our \textsc{DiffusionNER} can also decode entities in a non-autoregressive manner, and thus result in a faster inference speed with better performance.

\subsection{Diffusion Model}
Diffusion model is a deep latent generative model proposed by \citep{pmlr-v37-sohl-dickstein15}. With the development of recent work~\citep{ddpm}, diffusion model has achieved impressive results on image and audio generation \citep{rombach2021highresolution, dalle2, kong2021diffwave}. Diffusion model consists of the forward diffusion process and the reverse diffusion process. The former progressively disturbs the data distribution by adding noise with a fixed variance schedule \citep{ddpm}, and the latter learns to recover the data structure. Despite the success of the diffusion model in continuous state spaces (image or waveform), the application to natural language still remains some open challenges due to the discrete nature of text \citep{NEURIPS2021_958c5305, hoogeboom2022autoregressive, Self_conditioned_Embedding_Diffusion, DiffusionBERT}. Diffusion-LM \citep{Li-2022-DiffusionLM} models discrete text in continuous space through embedding and rounding operations and proposes an extra classifier as a guidance to impose constraints on controllable text generation. DiffuSeq \citep{gong2022diffuseq} and SeqDiffuSeq \citep{SeqDiffuSeq} extend diffusion-based text generation to a more generalized setting. They propose classifier-free sequence-to-sequence diffusion frameworks based on encoder-only and encoder-decoder architectures, respectively. 

Although diffusion models have shown their generative capability on images and audio, its potential on discriminative tasks has not been explored thoroughly.
Several pioneer works \citep{SegDiff, baranchuk2022labelefficient, diffusiondet} have made some attempts on diffusion models for object detection and semantic segmentation.
Our proposed \textsc{DiffusionNER} aims to  solve an extractive task on discrete text sequences.

\section{Preliminary}
\label{sec:3.1}

In diffusion models, both the forward and reverse processes can be considered a Markov chain with progressive Gaussian transitions. 
Formally, given a data distribution $\mathbf{x}_0 \sim q\left(\mathbf{x}_0\right)$ and a predefined variance schedule $\{\beta_{1}, \ldots, \beta_{T}\}$, the forward process $q$ gradually adds Gaussian noise with variance $\beta_t \in(0,1)$ at timestep $t$ to produce latent variables $\mathbf{x}_1, \mathbf{x}_2, \ldots, \mathbf{x}_T$ as follows:
\begin{align}
    &q\left(\mathbf{x}_1, \ldots, \mathbf{x}_T \mid \mathbf{x}_0\right) =\prod_{t=1}^T q\left(\mathbf{x}_t \mid \mathbf{x}_{t-1}\right) \\
q&\left(\mathbf{x}_t \mid \mathbf{x}_{t-1}\right)  =\mathcal{N}\left(\mathbf{x}_t; \sqrt{1-\beta_t} \mathbf{x}_{t-1}, \beta_t \mathbf{I}\right)
\end{align}
An important property of the forward process is that we can sample the noisy latents at an arbitrary timestep conditioned on the data $x_0$. With the notation $\alpha_t:=1-\beta_t$ and $\bar{\alpha}_t:=\prod_{s=0}^t \alpha_s$, we have:
\begin{align}
\label{eq:q_xt}
q\left(\mathbf{x}_t \mid \mathbf{x}_0\right) & =\mathcal{N}\left(\mathbf{x}_t ; \sqrt{\bar{\alpha}_t} \mathbf{x}_0,\left(1-\bar{\alpha}_t\right) \mathbf{I}\right)
\end{align}

\noindent As $\bar{\alpha}_T$ approximates 0, $\mathbf{x}_T$ follows the standard Gaussian distribution: $p\left(\mathbf{x}_T\right) \approx \mathcal{N}\left(\mathbf{x}_T; \mathbf{0}, \mathbf{I}\right)$. Unlike the fixed forward process, the \textit{reverse process} $p_\theta\left(\mathbf{x}_{0: T}\right)$ is defined as a Markov chain with learnable Gaussian transitions starting at a prior $p\left(\mathbf{x}_T\right)=\mathcal{N}\left(\mathbf{x}_T ; \mathbf{0}, \mathbf{I}\right)$:
\begin{align*}
p_\theta\left(\mathbf{x}_{0: T}\right)&=p\left(\mathbf{x}_T\right) \prod_{t=1}^T p_\theta\left(\mathbf{x}_{t-1} \mid \mathbf{x}_t\right) \\
p_\theta\left(\mathbf{x}_{t-1} \mid \mathbf{x}_t\right)&=\mathcal{N}\left(\mathbf{x}_{t-1} ; \mu_\theta\left(\mathbf{x}_t, t\right), \Sigma_\theta\left(\mathbf{x}_t, t\right)\right)
\end{align*}

\noindent where $\theta$ denotes the parameters of the model and ${\mu}_{\theta}$ and ${\Sigma}_\theta$ are the predicted covariance and mean of $q\left(\mathbf{x}_{t-1} \mid \mathbf{x}_t\right)$.
We set ${\Sigma}_\theta\left(\mathbf{x}_t, t\right)=\sigma_t^2 \mathbf{I}$ and build a neural network $f_\theta$ to predict the data $x_0$, denoted as $\hat{\mathbf{x}}_0=f_\theta\left(\mathbf{x}_t, t\right)$. Then we have ${\mu}_\theta\left(\mathbf{x}_t, t\right)=\tilde{{\mu}}_t\left(\mathbf{x}_t, \hat{\mathbf{x}}_0\right)=\tilde{{\mu}}_t\left(\mathbf{x}_t, f_\theta\left(\mathbf{x}_t, t\right)\right)$, where $\tilde{\mu}_t$ denotes the mean of posterior $q\left(\mathbf{x}_{t-1} \mid \mathbf{x}_t,\hat{\mathbf{x}}_0\right)$.
The reverse process is trained by optimizing a variational upper bound of $-\log \left(p_\theta\left(\mathbf{x}_{0}\right)\right)$. According to the derivation in \citet{ddpm}, we can simplify the training objective of the diffusion model by training the model $f_\theta(\cdot)$ to predict the data $\mathbf{x}_0$.

\section{Method}

In this section,  we first present the formulation of diffusion model for NER (i.e., the boundary denoising diffusion process) in \Cref{sec:3.2}. Then, we detail the architecture of the denoising network for boundary reverse process in \Cref{sec:3.3}. Finally, we describe the inference procedure of \textsc{DiffusionNER} in \Cref{sec:3.4}.

\begin{figure*}
    \centering
    \includegraphics[width=\linewidth]{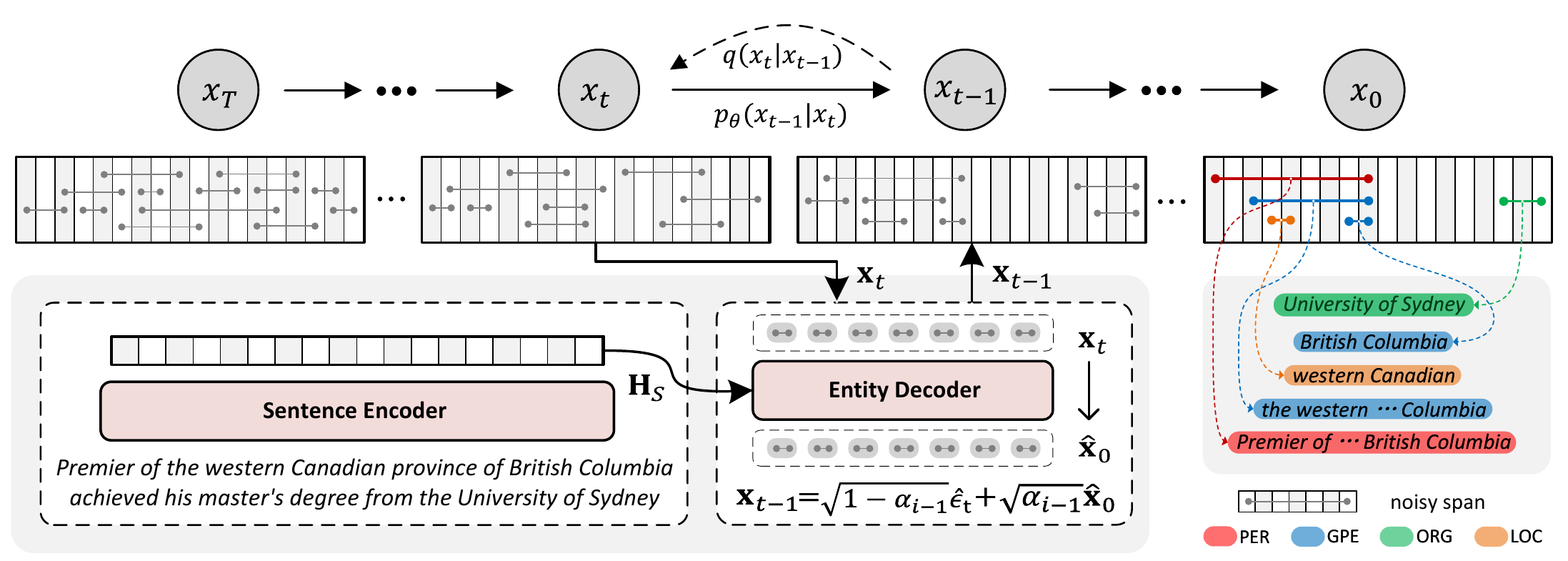}
    \caption{Overview of \textsc{DiffusionNER}. Boundary denoising diffusion process for NER with a denoising network.}
    \label{fig:overview}
\end{figure*}

\subsection{Boundary Denoising Diffusion Model}
\label{sec:3.2}

Given a sentence $S$ with length $M$, the named entity recognition task is to extract the entities $E = \{(l_i, r_i, t_i)\}_{i=0}^{N}$ contained in the sentence, where $N$ is the number of entities and $l_i, r_i, t_i$ denote the left and right boundary indices and type of the $i$-th entity. We formulate NER as a boundary denoising diffusion process, as shown in Figure~\ref{fig:overview}.
We regard entity boundaries as data samples, then the boundary forward diffusion is to add Gaussian noise to the entity boundaries while the reverse diffusion process is to progressively recover the original entity boundaries from the noisy spans.

\paragraph{Boundary Forward Diffusion} 

Boundary forward diffusion is the process of adding noise to the entity boundary in a stepwise manner.
In order to align the number of entities in different instances, we first expand the entity set to a fixed number $K$ ($>N$).
There are two ways to expand the entities, \textit{repetition strategy} and \textit{random strategy}, which add $K-N$ entities by duplicating entities or sampling random spans from a Gaussian distribution\footnote{\;We will discuss these two practices in \Cref{ana:expansion}.}.
For convenience, we use $\mathbf{B}\in\mathbb{R}^{K\times 2}$ to denote the boundaries of the $K$ expanded entities, with all of them normalized by the sentence length $M$ and scaled to $(-\lambda, \lambda)$ interval.

 Formally, given the entity boundaries as data samples $\mathbf{x}_0=\mathbf{B}$, we can obtain the noisy spans at timestep $t$ using the forward diffusion process. According to ~\Cref{eq:q_xt}, we have:
\begin{align}
\mathbf{x}_t & =\sqrt{\bar{\alpha}_t} \mathbf{x}_0+\sqrt{1-\bar{\alpha}_t } \epsilon
\end{align}

\noindent where $\epsilon \sim \mathcal{N}(\textbf{0}, \mathbf{I})$ is the noise sampled from the standard Gaussian. At each timestep, the noisy spans have the same shape as $\mathbf{x}_0$, i.e., $\mathbf{x}_1, \mathbf{x}_2, \ldots, \mathbf{x}_T  \in \mathbb{R}^{K\times 2}$.

\paragraph{Boundary Reverse Diffusion}
Starting from the noisy spans $\mathbf{x}_T$ sampled from the Gaussian distribution, boundary reverse diffusion adopts a non-Markovian denoising practice used in DDIM \citep{song2021denoising} to recover entities boundaries. Assuming $\tau$ is an arithmetic subsequence of the complete timestep sequence $[1,\ldots, T]$ of length $\gamma$ with $\tau_\gamma=T$. Then we refine the noisy spans $\mathbf{x}_{\tau_i}$ to $\mathbf{x}_{\tau_{i-1}}$ as follows:
\begin{align}
\label{eq:x0}
\mathbf{\hat{x}}_0 &= f_\theta(\mathbf{x}_{\tau_i}, S, \tau_i)  \\
\mathbf{\hat{\epsilon}}_{\tau_i} &= \frac{\mathbf{x}_{\tau_i}-\sqrt{\alpha_{\tau_i}} \mathbf{\hat{x}}_0}{\sqrt{1-\alpha_{\tau_i}}} \\
\mathbf{x}_{\tau_{i-1}} =& \sqrt{\alpha_{\tau_{i-1}}} \mathbf{\hat{x}}_0+\sqrt{1-\alpha_{\tau_{i-1}}} \mathbf{\hat{\epsilon}}_{\tau_i}
\label{eq:xtau-1}
\end{align}
\noindent where $\mathbf{\hat{x}}_0$ and $\mathbf{\hat{\epsilon}}_{\tau_i}$ are the predicted entity boundary and noise at timestep $\tau_i$. $f_{\theta}(\mathbf{x}_t, S, t)$ is a learnable denoising network and we will cover the network architecture in the next section (\Cref{sec:3.3}). After $\gamma$ iterations of DDIM, the noisy spans are progressively refined to the entity boundaries.

\subsection{Network Architecture}
\label{sec:3.3}
Denoising network $f_\theta(\mathbf{x}_{t}, S, t)$ accepts the noisy spans $\mathbf{x}_{t}$ and the sentence $S$ as inputs and predicts the corresponding entity boundaries $\mathbf{\hat{x}}_0$. As shown in Figure~\ref{fig:overview}, we parameterize the denoising network with a sentence encoder and an entity decoder.

\paragraph{Sentence Encoder} consists of a BERT \citep{devlin-etal-2019-bert} plus a stacked bi-directional LSTM. The whole span encoder takes the sentence ${S}$ as input and outputs the sentence encoding $\mathbf{H}_S \in \mathbb{R}^{M\times h}$. The sentence encoding $\mathbf{H}_S$ will be calculated only once and reused across all timesteps to save computations.


\paragraph{Entity Decoder} uses the sentence encoding $\mathbf{H}_S$ to first compute the representations of $K$ noisy spans $\mathbf{x}_t$ and then predicts the corresponding entity boundaries.
Specifically, we discretize the noisy spans into word indexes by rescaling, multiplying and rounding\footnote{\;First scaled with $\frac{1}{\lambda}$, then multiplied by $M$, and finally rounded to integers.}, then perform mean pooling over the inner-span tokens. The extracted span representations can be denoted as $\mathbf{H}_{{X}} \in \mathbb{R}^{K\times h}$. 
To further encode the spans, we design a span encoder that consists of a self-attention and a cross-attention layer. The former enhances the interaction between spans with key, query, and value as $\mathbf{H}_{{X}}$.  And the latter fuses the sentence encoding to the span representation with key, value as $\mathbf{H}_S$, and query as $\mathbf{H}_{{X}}$. 
We further add the sinusoidal embedding $\mathbf{E}_t$ \citep{NIPS2017_3f5ee243} of timestep $t$ to the span representations. 
Thus the new representations $\bar{\mathbf{H}}_{{X}}$ of the noisy spans can be computed:
\begin{align*}
        \bar{\mathbf{H}}_{{X}}&=\operatorname{SpanEncoder}(\mathbf{H}_S, \mathbf{H}_{{X}}) + \mathbf{E}_t ,
\end{align*}
Then we use two boundary pointers to predict the entity boundaries. For boundary $\delta\in\{l, r\}$, we compute the fusion representation $\mathbf{H}^{\delta}_{SX} \in \mathbb{R}^{K\times M \times h}$ of the noisy spans and the words, and then the probability of the word as the left or right boundaries $\mathbf{P}^{\delta}\in \mathbb{R}^{K\times M}$ can be computed as:
\begin{align*}
    \mathbf{H}^{\delta}_{SX}&=\mathbf{H}_S \mathbf{W}^{\delta}_S+\bar{\mathbf{H}}_{X} \mathbf{W}^{\delta}_X \\
    \mathbf{P}^{\delta}&=\operatorname{sigmoid}(\operatorname{MLP}(\mathbf{H}^{\delta}_{SX}))
\end{align*}
where $\mathbf{W}^{\delta}_S,\mathbf{W}^{\delta}_X \in \mathbb{R}^{h\times h}$ are two learnable matrixes and MLP is a two-layer perceptron. Based on the boundary probabilities, we can predict the boundary indices of the $K$ noisy spans.
If the current step is not the last denoising step, we compute $\mathbf{\hat{x}}_0$ by normalizing the indices with sentence length $M$ and scaling to $(-\lambda, \lambda)$ intervals. Then we conduct the next iteration of the reverse diffusion process according to \Crefrange{eq:x0}{eq:xtau-1}.

\begin{algorithm}[t]
\small
 \caption{Training}
 \label{alg:training}
 \Repeat{\normalfont{converged}}{
  Sample a sentence $S$ with entities $E$ from $\mathcal{D}$ \\
  Expand $E$ and get entity boundaries $\textbf{B}$ \\
  $\mathbf{x}_0 = \textbf{B} \in \mathbb{R}^{K\times 2}$\\
  $t \sim$ Uniform $(\{1, \ldots, T\})$ \\
  $\epsilon \sim \mathcal{N}(\mathbf{0}, \mathbf{I})$ \\
  $\mathbf{x}_t =\sqrt{\bar{\alpha}_t} \mathbf{x}_0+\sqrt{1-\bar{\alpha}_t } \epsilon$ \\
  Compute $\mathbf{P}^l$, $\mathbf{P}^r$ and $\mathbf{P}^c$  by running $f_{\theta}(\mathbf{x}_t, S, t)$ \\
  Take gradient descent step by optimize $-\sum_{i=1}^K\left(\log \mathbf{P}_i^{c}(\pi^c(i))+\sum_{\delta \in{l, r}} \log \mathbf{P}_i^{\delta}(\pi^\delta(i))\right)$
  
 }
\end{algorithm}

It is worth noting that we should not only locate entities but also classify them in named entity recognition. Therefore, we use an entity classifier to classify the noisy spans. The classification probability $\mathbf{P}^{c} \in \mathbb{R}^{K\times \mathcal{C}}$ is calculated as follows:
\begin{align*}
    \mathbf{P}^{c}=\operatorname{Classifier}(\bar{\mathbf{H}}_{X})
\end{align*}
\noindent where $\mathcal{C}$ is the number of entity types and Classifier is a two-layer perceptron with a softmax layer.

\begin{algorithm}[t]
\small
 \caption{Inference}
 \label{alg:inference}
 $\mathbf{x}_T \sim \mathcal{N}(\mathbf{0}, \mathbf{I}) \in \mathbb{R}^{K_{eval}\times 2}$ \\
 $\tau$ is an arithmetic sequence of length $\gamma$ with $\tau_\gamma=T$\\
 \For{$i = \gamma, \ldots, 1$}{
    Compute $\hat{\mathbf{x}}_0$, $\mathbf{P}^l$, $\mathbf{P}^r$ and $\mathbf{P}^c$ via $f_{\theta}(\mathbf{x}_t, S, t)$ \\
    $\mathbf{x}_{\tau_{i-1}} = \sqrt{\alpha_{\tau_{i-1}}} \mathbf{\hat{x}}_0+\sqrt{1-\alpha_{\tau_{i-1}}} \cdot \frac{\mathbf{x}_{\tau_i}-\sqrt{\alpha_{\tau_i}} \mathbf{\hat{x}}_0}{\sqrt{1-\alpha_{\tau_i}}}$
 }
 Decode entities ${(l_i, r_i, c_i)}^{K_{eval}}_{i=0}$, where $\delta_i = \operatorname{argmax}\mathbf{P}_i^\delta, \delta \in \{l, r, c\}$ \\
 Perform post-processing on ${(l_i, r_i, c_i)}^{K_{eval}}_{i=0}$ \\
 \Return final entities
\end{algorithm}

\paragraph{Training Objective} 
With $K$ entities predicted from the noisy spans and $N$ ground-truth entities, we first use the Hungarian algorithm \citep{kuhn1955hungarian} to solve the optimal matching $\hat{\pi}$ between the two sets\footnote{\;See \Cref{app:pi} for the solution of the optimal match $\hat{\pi}$.} as in \citet{detr}. $\hat{\pi}(i)$ denotes the ground-truth entity corresponding to the $i$-th noisy span. Then, we train the boundary reverse process by maximizing the likelihood of the prediction:
\begin{align*}
\mathcal{L}=-\sum_{i=1}^K\sum_{\delta \in\{l, r, c\}} \log \mathbf{P}_i^{\delta}\left(\hat{\pi}^\delta(i)\right)
\end{align*}

\noindent where $\hat{\pi}^l(i)$, $\hat{\pi}^r(i)$ and $\hat{\pi}^c(i)$ denote the left and right boundary indexes and type of the $\hat{\pi}(i)$ entity. Overall, \Cref{alg:training} displays the whole training procedure of our model for an explanation.

\begin{table*}[]
\centering
\small
\begin{tabular}{lcccccccccccc}
\toprule
\multicolumn{1}{c}{\multirow{2}{*}{Model}}   & \multicolumn{3}{c}{ACE04} & \multicolumn{3}{c}{ACE05}   & \multicolumn{3}{c}{GENIA} & \multirow{2}{*}{\makecell[c]{Agerage\\F1-score}}\\

 \cmidrule(lr){2-4}  \cmidrule(lr){5-7} \cmidrule(lr){8-10} 
&  Pr.  & Rec. & F1 & Pr.  & Rec. & F1  & Pr.  & Rec. & F1  \\
\midrule
\multicolumn{3}{l}{\textbf{Tagging-based}} \\
\citet{strakova-etal-2019-neural} & - & - & 81.48   & - & - & 80.82 & - & - & 77.80  & 80.03\\
\citet{ju-etal-2018-neural} & - & - & - & 74.20 & 70.30 & 72.20 & 78.50 & 71.30 & 74.70 & -\\
\citet{wang-etal-2020-pyramid} & 86.08 & 86.48 & 86.28 & 83.95 & 85.39 & 84.66 & 79.45 & 78.94 & 79.19 & {83.57}\\
\midrule
\multicolumn{3}{l}{\textbf{Generation-based}} \\
\citet{strakova-etal-2019-neural} & - & - & 84.40 & - & - & 84.33 & - & - & 78.31  & 82.35\\
\citet{yan-etal-2021-unified-generative} & 87.27 & 86.41 & 86.84 & 83.16 & 86.38 & 84.74 & 78.87 & 79.60 & 79.23  & {83.60} \\
 \citet{ijcai2021-542} & 88.46 & 86.10 & 87.26 & \textbf{87.48} & 86.63 & 87.05 & 82.31 & 78.66 & 80.44 & {84.91} \\
\citet{lu-etal-2022-unified} & - & - & 86.89 & - & - & 85.78 & - & - & - & -\\
\midrule
\multicolumn{3}{l}{\textbf{Span-based}} \\
\citet{yu-etal-2020-named}       & 87.30  & 86.00  & 86.70 & 85.20  & 85.60  & 85.40  & 81.80 & 79.30 & 80.50 & {84.20}\\
\citet{li-etal-2020-unified} & 85.05 & 86.32 &  85.98 & 87.16 & 86.59 & 86.88 & 81.14 & 76.82 & 78.92 &{83.92}\\
\citet{shen-etal-2021-locate} & 87.44 & 87.38 & 87.41 & 86.09 & 87.27 & 86.67 &80.19  &  80.89  &  80.54 & {84.87} \\
\citet{wan-etal-2022-nested} & 86.70  &  85.93  &  86.31  & 84.37  & 85.87 & 85.11   & 77.92 & 80.74 & 79.30 & {83.57}\\
 \citet{lou-etal-2022-nested} & 87.39 &  88.40 &  87.90 &  85.97 &  87.87 &  86.91  & - & - & - & -\\
  \citet{zhu-li-2022-boundary} & \textbf{88.43} & 87.53 &  87.98 & 86.25 &  88.07 &  \textbf{87.15} &  - &  - &  - & -\\
  \citet{yuan-etal-2022-fusing} & 87.13 &  87.68 &  87.40 &  86.70 &  86.94 &  86.82 & 80.42 & \textbf{82.06} & 81.23 & {85.14} \\
  \citet{li2022unified} & 87.33 &  87.71 &  87.52 & 85.03 & \textbf{88.62} &  86.79 & \textbf{83.10} & 79.76 & {81.39} & {85.23} \\
\midrule
\textbf{\textsc{DiffusionNER}} & 88.11 & \textbf{88.66} & \textbf{88.39} &  86.15 & 87.72 & 86.93 & 82.10 & 80.97 & \textbf{81.53} &\textbf{85.62}\\
\bottomrule
\end{tabular}
\caption{Results on {nested} NER datasets.}
\label{tab:nested}
\end{table*}

\subsection{Inference}
\label{sec:3.4}

During inference, \textsc{DiffusionNER} first samples $K_{eval}$ noisy spans from a Gaussian distribution, then performs iterative denoising with the learned boundary reverse diffusion process based on the denoising timestep sequence $\tau$.
Then with the predicted probabilities on the boundaries and type, we can decode $K_{eval}$ candidate entities ${(l_i, r_i, c_i)}^{K_{eval}}_{i=0}$, where $\delta_i = \operatorname{argmax}\mathbf{P}_i^\delta, \delta \in \{l, r, c\}$.
After that, we employ two simple post-processing operations on these candidates: de-duplication and filtering. For spans with identical boundaries, we keep the one with the maximum type probability. For spans with the sum of prediction probabilities less than the threshold $\varphi$, we discard them. The inference procedure is shown in \Cref{alg:inference}.

\section{Experimental Settings}

\subsection{Datasets}

For nested NER, we choose three widely used datasets for evaluation: ACE04~\citep{ doddington-etal-2004-automatic}, ACE05~\citep{2005-automatic}, and GENIA~\citep{10.5555/1289189.1289260}. ACE04 and ACE05 belong to the news domain and GENIA is in the biological domain. For flat NER, we use three common datasets to validate: CoNLL03~\citep{tjong-kim-sang-de-meulder-2003-introduction}, OntoNotes~\citep{pradhan-etal-2013-towards}, and MSRA~\citep{levow-2006-third}. More details about datasets can be found in \Cref{app:datasets}.

\subsection{Baselines}

We choose a variety of recent advanced methods as our baseline, which include: 1) Tagging-based methods~\citep{strakova-etal-2019-neural, ju-etal-2018-neural, wang-etal-2020-pyramid}; 2) Span-based methods~\citep{yu-etal-2020-named, li-etal-2020-unified, wan-etal-2022-nested, lou-etal-2022-nested, zhu-li-2022-boundary, yuan-etal-2022-fusing}; 3) Generation-based methods~\citep{ijcai2021-542, yan-etal-2021-unified-generative, lu-etal-2022-unified}. More details about baselines can be found in \Cref{app:baselines}.

\subsection{Implementation Details}
For a fair comparison, we use \texttt{bert-large}~\citep{devlin-etal-2019-bert} on ACE04, ACE05, CoNLL03 and
OntoNotes, \texttt{biobert-large}~\citep{chiu-etal-2016-train} on GENIA and \texttt{chinese-bert-wwm}~\citep{cui-etal-2020-revisiting} on MSRA. We adopt the Adam~\citep{adam} as the default optimizer with a linear warmup and linear decay learning rate schedule. The peak learning rate is set as $2e-5$ and the batch size is 8. For diffusion model, the number of noisy spans $K$ ($K_{eval}$) is set as 60, the timestep $T$ is 1000, and the sampling timestep $\gamma$ is 5 with a filtering threshold $\varphi=2.5$. The scale factor $\lambda$ for noisy spans is 1.0.
Please see \Cref{app:settings} for more details.

\section{Results and Analysis}

\subsection{Performance}

Table~\ref{tab:nested} illustrates the performance of \textsc{DiffusionNER} as well as baselines on the nested NER datasets. Our results in Table~\ref{tab:nested} demonstrate that \textsc{DiffusionNER} is a competitive NER method, achieving comparable or superior performance compared to state-of-the-art models on the nested NER. Specifically, on ACE04 and GENIA datasets, \textsc{DiffusionNER} achieves F1 scores of 88.39\% and 81.53\% respectively, with an improvement of +0.77\% and +0.41\%. And on ACE05, our method achieves comparable results. Meanwhile, \textsc{DiffusionNER} also shows excellent performance on flat NER, just as shown in Table 2.  We find that \textsc{DiffusionNER} outperforms the baselines on OntoNotes with +0.16\% improvement and achieves a comparable F1-score on both the English CoNLL03 and the Chinese MSRA. These improvements demonstrate that our \textsc{DiffusionNER} can locate entities more accurately due to the benefits of progressive boundary refinement, and thus obtain better performance. The results also validate that our \textsc{DiffusionNER} can recover entity boundaries from noisy spans via boundary denoising diffusion.

\begin{table}[]
\centering
\small
\begin{tabular}{lccc}
\toprule
\multicolumn{1}{c}{\multirow{2}{*}{Model}}   & \multicolumn{3}{c}{CoNLL03}  \\

 \cmidrule(lr){2-4} 
&  Pr.  & Rec. & F1 \\
\midrule
\citet{lu-etal-2022-unified} & - & -  & 92.99  \\
\citet{shen-etal-2021-locate} & 92.13 & {93.73} & 92.94  \\
\citet{li-etal-2020-unified}$^\dagger$ & 92.33 & \textbf{94.61} & 93.04 \\
\citet{yan-etal-2021-unified-generative} & 92.56 & 93.56 & 93.05  \\
\citet{li2022unified}$^\dagger$ & 92.71 & 93.44 & \textbf{93.07}  \\
\midrule
\textbf{\textsc{DiffusionNER}} & \textbf{92.99} & 92.56 & 92.78  \\
\toprule
\multicolumn{1}{c}{\multirow{2}{*}{Model}}    & \multicolumn{3}{c}{OntoNotes}    \\

 \cmidrule(lr){2-4}  
&  Pr.  & Rec. & F1   \\
\midrule
\citet{https://doi.org/10.48550/arxiv.1911.04474} & -  & -  & 89.78\\
\citet{yan-etal-2021-unified-generative}  & 89.62 & 90.92 & 90.27 \\
\citet{li-etal-2020-unified}$^\dagger$ & {90.14} & 89.95 & {90.02}  \\
\citet{li2022unified}$^\dagger$ &{90.03} & 90.97 & {90.50}  \\
\midrule
\textbf{\textsc{DiffusionNER}} & \textbf{90.31} & \textbf{91.02} & \textbf{90.66}  \\

\toprule
\multicolumn{1}{c}{\multirow{2}{*}{Model}}    & \multicolumn{3}{c}{MSRA}    \\

 \cmidrule(lr){2-4}  
&  Pr.  & Rec. & F1   \\
\midrule
\citet{https://doi.org/10.48550/arxiv.1911.04474} & -  & -  & 92.74  \\
\citet{shen-etal-2021-locate} &  92.20 & 90.72 & 91.46  \\
\citet{li-etal-2020-unified}$^\dagger$ & 91.98 & 93.29 & 92.63 \\

\citet{li2022unified}$^\dagger$ & {94.88} & \textbf{95.06} & {94.97} \\
\midrule
\textbf{\textsc{DiffusionNER}} & \textbf{95.71} & 94.11 & 94.91 \\
\bottomrule
\end{tabular}
\caption{Results on \textit{flat} NER datasets. $\dagger$ means that we reproduce the results under the same setting.}
\label{tab:flat}
\end{table}

\subsection{Analysis}
\label{ana}

\paragraph{Inference Efficiency}
To further validate whether our \textsc{DiffusionNER} requires more inference computations, we also conduct experiments to compare the inference efficiency between \textsc{DiffusionNER} and other generation-based models~\citep{lu-etal-2022-unified,yan-etal-2021-unified}. Just as shown in \Cref{tab:speedup}, we find that \textsc{DiffusionNER} could achieve better performance while maintaining a faster inference speed with minimal parameter scale.
Even with a denoising timestep of $\gamma = 10$, \textsc{DiffusionNER} is 18$\times$ and 3$\times$ faster than them. This is because \textsc{DiffusionNER} generates all entities in parallel within several denoising timesteps, which avoids generating the linearized entity sequence in an autoregressive manner. In addition, \textsc{DiffusionNER} shares sentence encoder across timesteps, which further accelerates inference speed.

\begin{table}[!htp]
    \centering
    \small
    \begin{tabular}{>{\centering\arraybackslash}p{2.74cm}>{\centering\arraybackslash}p{0.55cm}>{\centering\arraybackslash}p{0.55cm}>{\centering\arraybackslash}p{0.72cm}>{\centering\arraybackslash}p{1.05cm}}
    \toprule
    Model & \# P & F1  & Sents/s & SpeedUp \\
    \midrule
     \citet{lu-etal-2022-unified} & 849M  & 86.89  & 1.98 & 1.00$\times$  \\  
     \citet{yan-etal-2021-unified} & 408M & 86.84   & 13.75 & 6.94$\times$ \\ 
     \midrule
     \textbf{\textsc{DiffusionNER$_{[\tau=1]}$}} & 381M & 88.40  & \textbf{82.44} & \textbf{41.64$\times$}  \\  
     \textbf{\textsc{DiffusionNER$_{[\tau=5]}$}} & 381M & 88.53  & 57.08 & 28.83$\times$  \\ 
     \textbf{\textsc{DiffusionNER$_{[\tau=10]}$}}  & 381M & \textbf{88.57}  & 37.10 & 18.74$\times$  \\  
    \bottomrule
    \end{tabular}
    \caption{Comparison with generation-based methods in terms of parameters, performance, and inference speed. \# P means the number of parameters. All experiments are conducted on a single GeForce RTX 3090 with the same setting. The results are reported on ACE04.}
    \label{tab:speedup}
\end{table}

\begin{figure}[!htp]
  \centering
\includegraphics[width=1.0\linewidth]{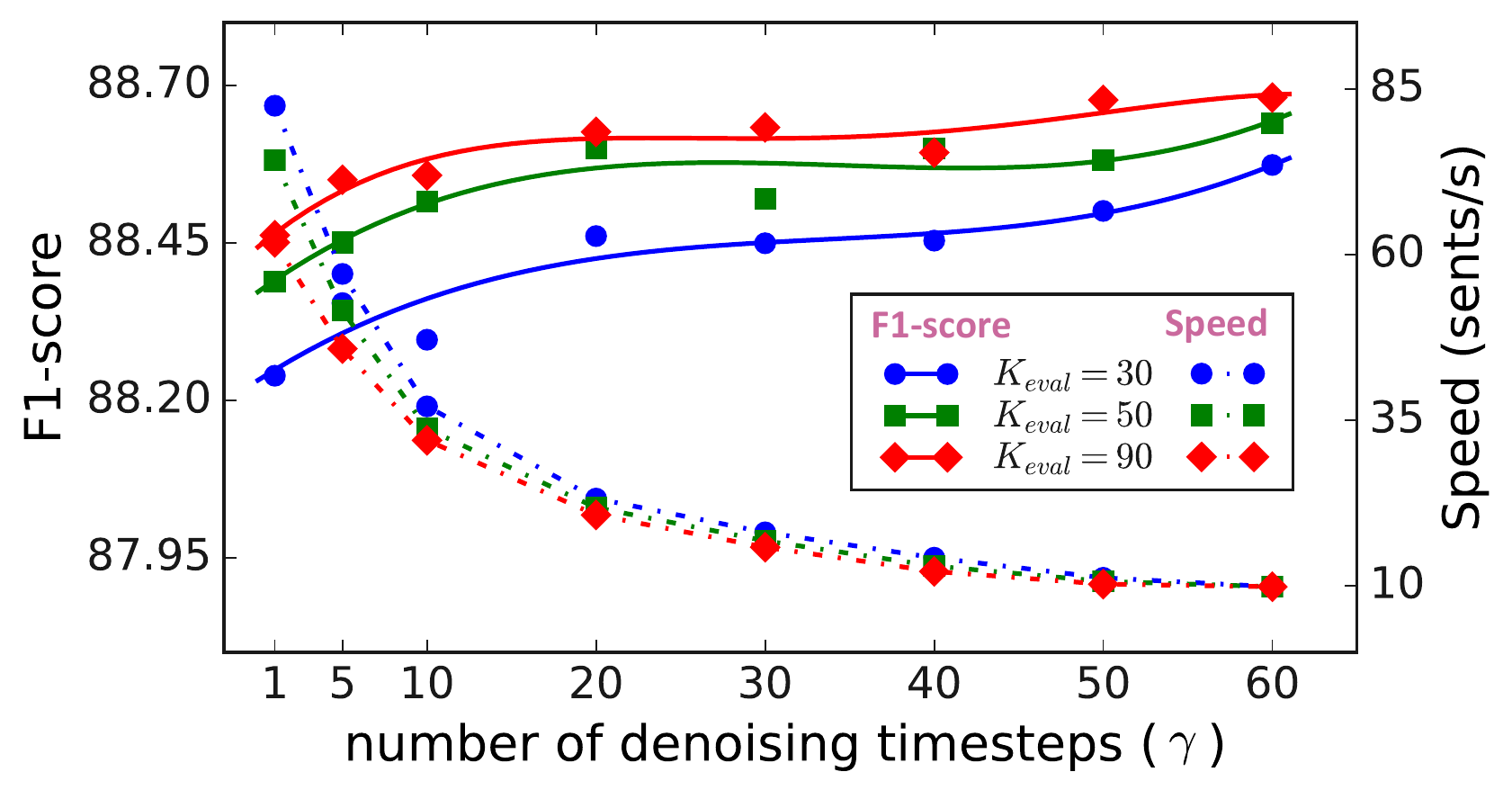}
  \caption{Analysis of denoising timestep $\gamma$ on ACE04.}
  \label{fig:gamma}
\end{figure}

\paragraph{Denoising Timesteps}
We also conduct experiments to analyze the effect of different denoising timesteps on model performance and inference speed of \textsc{DiffusionNER} under various numbers of noisy spans. Just as shown in \Cref{fig:gamma}, we find that, with an increase of denoising steps, the model obtains incremental performance improvement while sacrificing inference speed. Considering the trade-off between performance and efficiency, we set $\gamma = 5$ as the default setting. In addition, when the noisy spans are smaller, the improvement brought by increasing the denoising timesteps is more obvious. This study indicates that our DiffusionNER can effectively counterbalance the negative impact of undersampling noise spans on performance by utilizing additional timesteps.

\begin{figure}[!htp]
  \centering
  \includegraphics[width=0.91\linewidth]{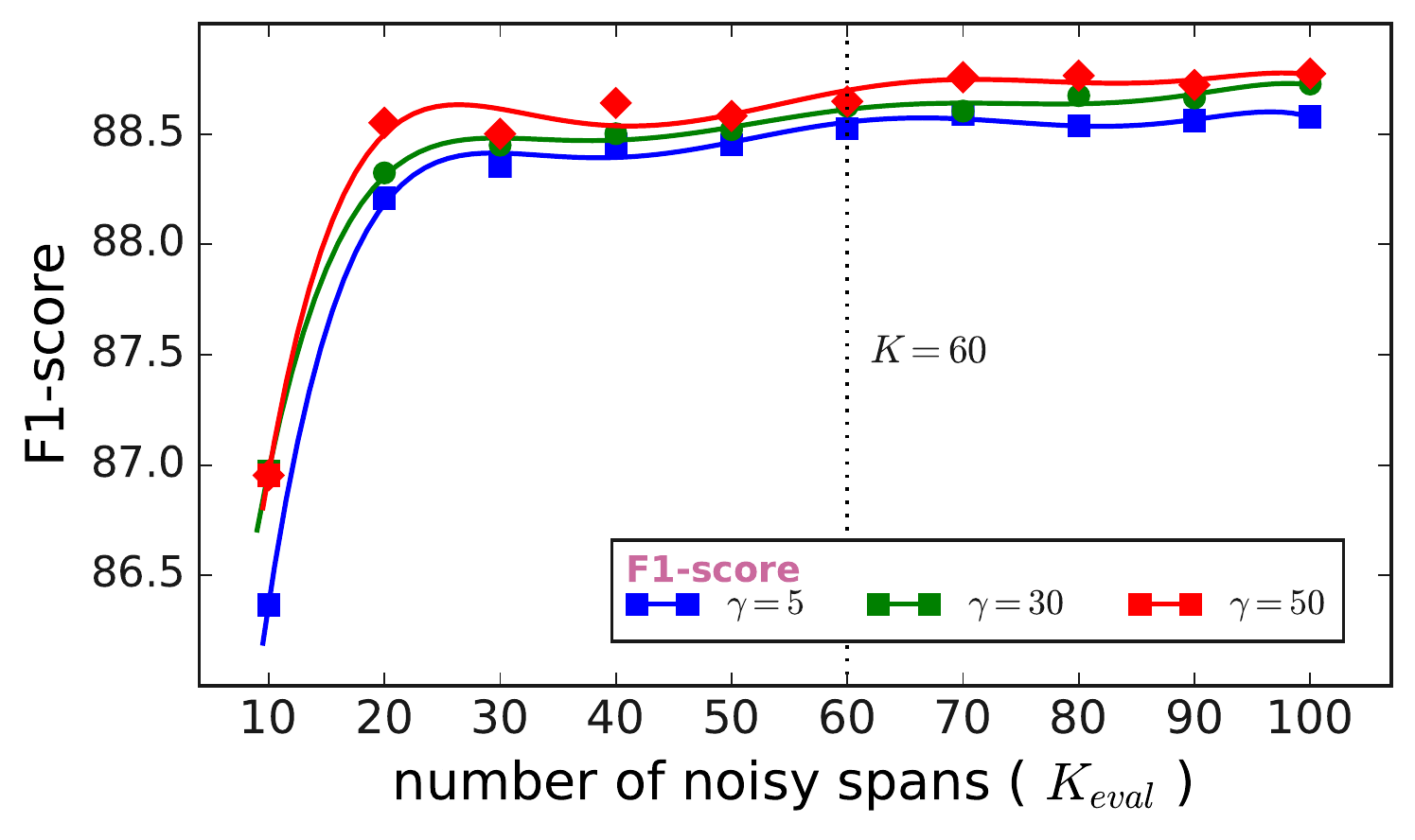}
  \caption{Analysis of \#sampled noisy spans on ACE04.}
  \label{fig:k}
\end{figure}

\paragraph{Sampling Number} 
As a generative latent model, \textsc{DiffusionNER} can decouple training and evaluation, and dynamically sample noisy spans during evaluation. To manifest this advantage, we train \textsc{DiffusionNER} on ACE04 with $K = 60$ noisy spans and evaluate it with different sampling numbers $K_{eval}$. The results are shown in \Cref{fig:k}. Overall, the model performance becomes better as the sampling number of noisy spans increases. Specifically, we find that \textsc{DiffusionNER} performs worse when $K_{eval}<30$. We guess this is because fewer noisy spans may not cover all potential entities. When sampling number $K_{eval}>60$, we find it could also slightly improve model performance. Overall, the dynamic sampling of noisy spans in \textsc{DiffusionNER} has the following advantages: 1) we can improve model performance by controlling it to sample more noisy spans; 2) dynamic sampling strategy also allows the model to predict an arbitrary number of entities in any real-world application, avoiding the limitations of the sampling number at the training stage.

\subsection{Ablation Study}
\label{app:ablationstudy}

\paragraph{Network Architecture}
\label{ana:pooling}
As shown in \Cref{tab:modelstructure}, we conduct experiments to investigate the network architecture of the boundary reverse diffusion process.
We found that \textsc{DiffusionNER} performs better with a stronger pre-trained language model (PLM), as evidenced by an improvement of +0.53\% on ACE04 and +0.11\% on CoNLL03 when using \texttt{roberta-large}.
Additionally, for the span encoder, we find that directly removing self-attention between noisy spans or cross-attention of spans to the sentence can significantly impair performance. When both are ablated, model performance decreases by 1.37\% and 1.15\% on ACE04 and CoNLL03. These results indicate that the interaction between the spans or noisy spans and the sentence is necessary.

\begin{table}[!htp]
    \centering
    \small
    \begin{tabular}{llccc}
    \toprule
    & Setting & ACE04 & CoNLL03\\
    \midrule
    \multirow{3}{*}{\rotatebox[origin=c]{90}{PLM}} &{RoBERTa-Large} & \textbf{88.99} & \textbf{92.89}\\
     &{BERT-Large} & 88.39 & {92.78}\\
     &{BERT-Base} & 86.93 & 92.02\\
     \midrule
    \multirow{4}{*}{\rotatebox[origin=c]{90}{Module}} &{\textsc{Default}} & \textbf{88.39} & \textbf{92.78} \\
    \cmidrule{2-4}
     &{w/o self-attention} & 87.94 & 92.25 \\
     &{w/o cross-attention} & 87.22 & 91.40 \\
     &{w/o span encoder} & 87.09 & 91.63 \\
    \bottomrule
    \end{tabular}
    \caption{Ablation study of network architecture.  }
    \label{tab:modelstructure}
\end{table}


\paragraph{Variance Scheduler}
The variance scheduler plays a crucial role in controlling the intensity of the added noise at each timestep during boundary forward diffusion process. Therefore, we analyze the performance of \textsc{DiffusionNER} on different variance schedulers with different noise timesteps $T$. The results on ACE04 and CoNLL03 are shown in \Cref{tab:variancescheduler}. We find that the cosine scheduler generally yields superior results on the ACE04, while the linear scheduler proves to be more effective on CoNLL03. In addition, the performance of \textsc{DiffusionNER} varies with the choice of noise timestep, with the best performance achieved at $T = 1000$ for ACE04 and $T = 1500$ for CoNLL03.

\begin{table}[]
    \centering
    \small
    \begin{tabular}{cccc}
    \toprule
    Scheduler & Timesteps ($T$) & ACE04 & CoNLL03\\
    \midrule
     \multirow{3}{*}{cosine} 
     &${T=1000}$ & \textbf{88.39}  & 91.56  \\
     &${T=1500}$ & 87.49 & \textbf{92.04}\\
     &${T=2000}$ & 88.33 & 91.79\\
    \midrule
    \multirow{3}{*}{linear} 
     &${T=1000}$ & \textbf{88.38}  & 92.78 \\
     &${T=1500}$ & 87.83 & \textbf{92.87} \\
     &${T=2000}$ & 88.17 & 92.56\\
    \bottomrule
    \end{tabular}
    \caption{Ablation study of variance scheduler. }
    \label{tab:variancescheduler}
\end{table}

\begin{table}[]
    \centering
    \small
    \begin{tabular}{cccc}
    \toprule
    Strategy & \# Noisy Spans& ACE04 & CoNLL03\\
    \midrule
     \multirow{3}{*}{Repetition} 
     &${K=\ \ 60}$ & 88.15  & 92.66 \\
     &${K=120}$ & \textbf{88.49} & 92.54 \\
     &${K=150}$ & 88.19 & \textbf{92.71} \\
    \midrule
    \multirow{3}{*}{Random} 
    & ${K=\ \ 60}$ & 88.46 & 92.78 \\
    & ${K=120}$ & \textbf{88.53} & \textbf{92.79} \\
    & ${K=150}$ & 88.11 & 92.60 \\
    \bottomrule
    \end{tabular}
    \caption{Ablation study of expansion strategy. }
    \label{tab:expansionstrategy}
\end{table}

\paragraph{Expansion Stratagy}
\label{ana:expansion}
The expansion stratagy of the entity set can make the number of $K$ noisy spans consistent across instances during training. We conduct experiments to analyze the performance of \textsc{DiffusionNER} for different expansion strategies with various numbers of noisy spans. The experimental results are shown in \Cref{tab:expansionstrategy}. Generally, we find that the random strategy could achieve similar or better performance than the repetitive strategy. In addition, \Cref{tab:expansionstrategy} shows that \textsc{DiffusionNER} is insensitive to the number of noisy spans during training. Considering that using more noisy spans brings more computation and memory usage, we set $K=60$ as the default setting.

\section{Conclusion}

In this paper, we present \textsc{DiffusionNER}, a novel generative approach for NER that converts the task into a boundary denoising diffusion process. Our evaluations on six nested and flat NER datasets show that \textsc{DiffusionNER} achieves comparable or better performance compared to previous state-of-the-art models. Additionally, our additional analyses reveal the advantages of \textsc{DiffusionNER} in terms of inference speed, progressive boundary refinement, and dynamic entity sampling. Overall, this study is a pioneering effort of diffusion models for extractive tasks on discrete text sequences, and we hope it may serve as a catalyst for more research about the potential of diffusion models in natural language understanding tasks.

\section*{Limitations}

We discuss here the limitations of the proposed \textsc{DiffusionNER}. First, as a latent generative model, \textsc{DiffusionNER} relies on sampling from a Gaussian distribution to produce noisy spans, which leads to a random characteristic of entity generation. Second, \textsc{DiffusionNER} converges slowly due to the denoising training and matching-based loss over a large noise timestep. Finally, since discontinuous named entities often contain multiple fragments, \textsc{DiffusionNER} currently lacks the ability to generate such entities.
We can design a simple classifier on top of \textsc{DiffusionNER}, which is used to combine entity fragments and thus solve the problem of discontinuous NER.

\section*{Acknowledgments}
This work is supported by the Key Research and Development Program of Zhejiang Province, China (No. 2023C01152),  the Fundamental Research Funds for the Central Universities (No. 226-2023-00060), and MOE Engineering Research Center of Digital Library.

\bibliography{anthology,custom}
\bibliographystyle{acl_natbib}

\clearpage
\appendix

\section{Optimal Matching $\hat{\pi}$}
\label{app:pi}

Given a fixed-size set of $K$ noisy spans, \textsc{DiffusionNER} infers $K$ predictions, where $K$ is larger than the number of $N$
entities in a sentence. One of the main difficulties of training is to assign the ground truth to the prediction. Thus we first produce an optimal bipartite matching between predicted and ground truth entities and then optimize the likelihood-based loss.

Assuming that $\hat{Y} = \{\hat{Y}_i\}^K_{i=1}$ are the set of $K$ predictions, where $\hat{Y}_i=\left(\mathbf{P}_i^l, \mathbf{P}_i^r, \mathbf{P}_i^c\right)$. We denote the ground truth set of $N$ entities as $Y=\left\{\left(l_i, r_i, c_i\right)\right\}_{i=1}^N$, where $l_i, r_i, c_i$ are the boundary indices and type for the $i$-th entity. Since $K$ is larger than the number of $N$ entities, we pad $Y$ with $\varnothing$ (no entity). To find
a bipartite matching between these two sets we search for a permutation of $K$ elements $\pi \in \mathfrak{S}(K)$ with the lowest cost:
\begin{equation*}
\hat{\pi}=\underset{\pi \in \mathfrak{S}(K)}{\arg \min } \sum_i^K \mathcal{L}_{\operatorname{match}}\left(\hat{Y}_i, {Y}_{\pi(i)}\right)
\end{equation*}
\noindent where $\mathcal{L}_{\operatorname{match}}\left(\hat{Y}_i, {Y}_{\pi(i)}\right)$ is a pair-wise matching cost between the prediction $\hat{Y}_i$ and ground truth ${Y}_{\pi(i)}$ with index $\pi(i)$. We define it as
$-\mathds{1}({Y}_{\pi(i)} \neq \varnothing)\sum_{\sigma\in \{l,r,c\}} \mathbf{P}_{i}^\sigma\left(Y^\sigma_{\pi(i)}\right)$,
\noindent where $\mathds{1}(\cdot)$ denotes an indicator function. Finally, the optimal assignment $\hat{\pi}$ can be computed with the Hungarian algorithm.

\begin{table*}[!ht]
\centering
\small
\begin{tabular}{l>{\centering\arraybackslash}p{0.9cm}>{\centering\arraybackslash}p{0.9cm}>{\centering\arraybackslash}p{0.9cm}>{\centering\arraybackslash}p{0.9cm}>{\centering\arraybackslash}p{0.9cm}>{\centering\arraybackslash}p{0.9cm}>{\centering\arraybackslash}p{0.9cm}>{\centering\arraybackslash}p{0.9cm}>{\centering\arraybackslash}p{0.9cm}}
\toprule
\multirow{2}{*}{}   & \multicolumn{3}{c}{ACE04}& \multicolumn{3}{c}{ACE05} & \multicolumn{2}{c}{GENIA} \\
 \cmidrule(lr){2-4}  \cmidrule(lr){5-7} \cmidrule(lr){8-9} 
& Train  & Dev & Test & Train  & Dev & Test  & Train   & Test   \\
\midrule
number of sentences &  6200 &  745 &  812 &  7194 &  969 &  1047 &    16692 &   1854  \\
\quad - with nested entities  &  2712 &  294 &  388 &  2691 &  338 &  320  &  3522 &   446  \\
number of entities &  22204 &  2514 &  3035 &  24441 &  3200 &  2993 &  50509 &    5506   \\
\quad - nested entities &  10149 &  1092 & 1417  & 9389 &  1112 &  1118 &  9064 &    1199 \\
\quad - nesting ratio (\%) &  45.71 & 46.69 &  45.61 & 38.41 & 34.75 &  37.35 &    17.95 &    21.78\\
average sentence length &  22.50 &  23.02 &  23.05 &  19.21 &  18.93 &  17.2  &  25.35 &    25.99 \\
maximum number of entities &  28 & 22 &  20 & 27 & 23 &  17 &  25 &  14  \\
average number of entities &  3.58 & 3.37 &  3.73 & 3.39 & 3.30 &  2.86  &  3.03 & 2.97 \\
\bottomrule
\end{tabular}

\begin{tabular}{lccccccccc}
\toprule
\multirow{2}{*}{}   & \multicolumn{3}{c}{CoNLL03}& \multicolumn{3}{c}{OntoNotes} & 
 \multicolumn{3}{c}{Chinese MSRA}\\
 \cmidrule(lr){2-4}  \cmidrule(lr){5-7} \cmidrule(lr){8-10}  
& Train  & Dev & Test & Train  & Dev & Test & Train  & Dev & Test   \\
\midrule
number of sentences  &  14041 &  3250 &  3453 &  49706 &  13900 &  10348  & 41728 & 4636 & 4365 \\
number of entities &  23499 &  5942 &  5648 &  128738 &  20354 &  12586   & 70446 & 4257 & 6181 \\
average sentence length &  14.50 &  15.80 &  13.45 & 24.94 &  20.11 &  19.74  & 46.87 & 46.17 & 39.54 \\
maximum number of entities &  20 & 20 &  31 & 32 & 71 &  21 & 125 & 18 & 461 \\
average number of entities &  1.67 & 1.83 &  1.64 & 2.59 & 1.46 &  1.22   & 1.69 & 0.92 & 1.42\\
\bottomrule

\end{tabular}

\caption{Statistics of the \textbf{\textit{nested}} 
 and \textbf{\textit{flat}} datasets used in our experiments.}
\label{tab:statistics}
\end{table*}

\section{Datasets}
\label{app:datasets}

We conduct experiments on six widely used NER datasets, including three nested and three flat datasets. Table \ref{tab:statistics} reports detailed statistics about the datasets.

\paragraph{ACE04 and ACE05} \citep{doddington-etal-2004-automatic, 2005-automatic} are two nested NER datasets and contain 7 entity categories, including \texttt{PER}, 
 \texttt{ORG}, \texttt{LOC}, \texttt{GPE}, \texttt{WEA},  \texttt{FAC} and \texttt{VEH} categories. We follow the same setup as previous works \citet{katiyar-cardie-2018-nested, lin-etal-2019-sequence}.

\paragraph{GENIA} \citep{10.5555/1289189.1289260} is a biology nested NER dataset and contains 5 entity types, including \texttt{DNA}, \texttt{RNA}, \texttt{protein}, \texttt{cell line} and \texttt{cell type} categories. Follow \citet{huang-etal-2022-pyramid, shen-etal-2021-locate}, we train the model on the concatenation of the train and dev sets.

\paragraph{CoNLL03}  \citep{tjong-kim-sang-de-meulder-2003-introduction} is a flat dataset with 4 types of named entities: \texttt{LOC}, \texttt{ORG}, \texttt{PER} and \texttt{MISC}. Follow \citet{yu-etal-2020-named, yan2021bartner, shen-etal-2021-locate}, we train our model on the combination of the train and dev sets.

\paragraph{OntoNotes} \citep{pradhan-etal-2013-towards} is a flat dataset with 18 types of named entities, including 11 entity types and 7 value types. We use the same train, development, and test splits as \citet{li-etal-2020-unified, shen-etal-2022-parallel}.

\paragraph{MSRA} \citep{levow-2006-third} is a Chinese flat dataset with 3 entity types, including \texttt{ORG}, \texttt{PER}, \texttt{LOC}. We keep the same dataset splits and pre-processing with \citet{li2022unified, shen-etal-2021-locate}.

\section{Detailed Parameter Settings}
\label{app:settings}

Entity boundaries are predicted at the word level, and we use max-pooling to aggregate subwords into word representations. We use the multi-headed attention with 8 heads in the span encoder, and add a feedforward network layer after the self-attention and cross-attention layer. During training, we first fix the parameters of BERT and train the model for $5$ epochs to warm up the parameters of the entity decoder. We tune the learning rate from $\{1e-5, 2e-5, 3e-5\}$ and the threshold $\varphi$ from range $[2.5, 2.7]$ with a step $0.05$, and select the best hyperparameter setting according to the performance of the development set. The detailed parameter settings are shown in \Cref{tb:hp}.

\begin{table}[!t]
\small
\centering
   \aboverulesep=0ex 
   \belowrulesep=0ex 
\renewcommand{\arraystretch}{1.3} 
\begin{tabular}{l|>{\centering\arraybackslash}p{1.1cm}>{\centering\arraybackslash}p{1.7cm}>{\centering\arraybackslash}p{0.8cm}}
\toprule
\textbf{Hyperparameter}  & \textbf{ACE04}& \textbf{ACE05}& \textbf{GENIA} \\
\hline
learning rate & 2e-5 & 3e-5 & 2e-5 \\
\hline
weight decay & 0.1 & 0.1 & 0.1 \\
\hline
lr warmup   & 0.1 & 0.1 & 0.1  \\
\hline
batch size & 8 & 8 & 8  \\
\hline
epoch & 100 & 50 & 50  \\
\hline
hidden size $h$ & 1024 & 1024 & 1024  \\
\hline
threshold $\varphi$ & 2.55 & 2.65 & 2.50 \\
\hline
scale factor $\lambda$  & 1.0 & 1.0 & 2.0  \\
\bottomrule
\end{tabular}

\begin{tabular}{l|>{\centering\arraybackslash}p{1.1cm}>{\centering\arraybackslash}p{1.7cm}>{\centering\arraybackslash}p{0.8cm}}
\toprule
\textbf{Hyperparameter}  & \textbf{CoNLL03}& \textbf{Ontonotes}& \textbf{MSRA} \\
\hline
learning rate & 2e-5 & 2e-5 & 5e-6 \\
\hline
weight decay & 0.1 & 0.1 & 0.1 \\
\hline
lr warmup   & 0.1 & 0.1 & 0.1  \\
\hline
batch size & 8 & 8 & 16  \\
\hline
epoch & 100 & 50 & 100  \\
\hline
hidden size $h$ & 1024 & 1024 & 768  \\
\hline
threshold $\varphi$ & 2.50 & 2.55 & 2.60 \\
\hline
scale factor $\lambda$  & 1.0 & 2.0 & 1.0  \\
\bottomrule
\end{tabular}

\caption{Detailed Hyperparameter Settings}
\label{tb:hp}
\end{table}

\section{Baselines}
\label{app:baselines}

We use the following models as baselines:

\begin{itemize}
    \item \textbf{LinearedCRF} \citep{strakova-etal-2019-neural} concatenates the nested
entity multiple labels into one multilabel, and uses CRF-based tagger to decode flat or nested entities.
    \item \textbf{CascadedCRF} \citep{ju-etal-2018-neural} stacks the flat NER layers and identifies nested entities in an inside-to-outside way.
    \item \textbf{Pyramid} \citep{wang-etal-2020-pyramid} constructs the representations of mentions from the bottom up by stacking flat NER layers in a pyramid, and allows bidirectional interaction between layers by an inverse pyramid.
    \item \textbf{Seq2seq} \citep{strakova-etal-2019-neural} converts the labels of nested entities into a sequence and then uses a seq2seq model to decode entities.
    \item \textbf{BARTNER} \citep{yan-etal-2021-unified-generative} is also a sequence-to-sequence framework that transforms entity labels into word index sequences and decodes entities in a word-pointer manner.
    \item \textbf{Seq2Set} \citep{ijcai2021-542}treats NER as a sequence-to-set task and constructs learnable entity queries to generate entities.
    \item \textbf{UIE} \citep{lu-etal-2022-unified} designs a special schema for the conversion of structured information to sequences, and adopts a generative model to generate linearized sequences to unify various information extraction tasks.
    \item \textbf{Biaffine} \citep{yu-etal-2020-named} reformulates NER as a structured prediction task and adopts a dependency parsing approach for NER.
    \item \textbf{MRC} \citep{li-etal-2020-unified} reformulates NER as a reading comprehension task and extracts entities to answer the type-specific questions.
    \item \textbf{Locate\&label} \citep{shen-etal-2021-locate} is a two-stage method that first regresses boundaries to locate entities and then performs entity typing.
    \item \textbf{SpanGraph} \citep{wan-etal-2022-nested} utilizes a retrieval-based span-level graph to improve the span representation, which can connect spans and entities in the training data.
    \item \textbf{LLCP} \citep{lou-etal-2022-nested} treat NER as latent lexicalized constituency parsing and resort to constituency trees to model nested entities.
    \item \textbf{BoundarySmooth} \citep{zhu-li-2022-boundary}, inspired by label smoothing, proposes boundary smoothing for span-based NER methods.
    \item \textbf{Triffine} \citep{yuan-etal-2022-fusing} proposes a triaffine mechanism to integrate heterogeneous factors to enhance the span representation, including inside tokens, boundaries, labels, and related spans.
    \item \textbf{Word2Word} \citep{li2022unified} treats NER as word-word relation classification and uses multi-granularity 2D convolutions to construct the 2D word-word grid representations.
\end{itemize}

\end{document}